\documentclass[sigconf,natbib=true,anonymous=false]{acmart}

\usepackage{multirow}
\usepackage{amsmath}
\usepackage{subfig} 
\usepackage[flushleft]{threeparttable}
\usepackage{makecell} 
\usepackage{enumitem} 
\usepackage{adjustbox} 
\usepackage{graphicx} 
\usepackage{subfig} 
\usepackage{pifont} 
\usepackage{makecell}

\usepackage{array}
\newcommand{\PreserveBackslash}[1]{\let\temp=\\#1\let\\=\temp}
\newcolumntype{C}[1]{>{\PreserveBackslash\centering}p{#1}}
\newcolumntype{R}[1]{>{\PreserveBackslash\raggedleft}p{#1}}
\newcolumntype{L}[1]{>{\PreserveBackslash\raggedright}p{#1}}

\makeatletter
\newcommand\footnoteref[1]{\protected@xdef\@thefnmark{\ref{#1}}\@footnotemark}
\makeatother

\DeclareMathOperator*{\E}{\mathbb{E}}

\AtBeginDocument{%
  }

\acmConference[SIGIR' 24]{The 47th International ACM SIGIR Conference on Research and Development in Information Retrieval}{July 14 - 18, 2024}{Washington D.C., USA}

\begin{document}

\title{Capability-aware Prompt Reformulation Learning for Text-to-Image Generation}


\author{Jingtao Zhan}
\email{jingtaozhan@gmail.com}
\affiliation{%
  \institution{Department of Computer Science and Technology, Tsinghua University}
  \institution{Quan Cheng Laboratory}
  \state{Beijing 100084}
  \country{China}
  \postcode{}
}

\author{Qingyao Ai}
\email{aiqy@tsinghua.edu.cn}
\authornote{Corresponding author}
\affiliation{%
  \institution{Quan Cheng Laboratory}
  \institution{Department of Computer Science and Technology, Tsinghua University}
  \state{Beijing 100084}
  \country{China}
}

\author{Yiqun Liu}
\email{yiqunliu@tsinghua.edu.cn}
\affiliation{%
  \institution{Department of Computer Science and Technology, Tsinghua University}
  \institution{Zhongguancun Laboratory}
  \state{Beijing 100084}
  \country{China}
}

\author{Jia Chen}
\email{chenjia2@xiaohongshu.com}
\affiliation{%
 \institution{Xiaohongshu Inc}
 \state{Beijing}
 \country{China}}

\author{Shaoping Ma}
\email{msp@tsinghua.edu.cn}
\affiliation{%
  \institution{Department of Computer Science and Technology, Tsinghua University}
  \institution{Zhongguancun Laboratory}
  \state{Beijing 100084}
  \country{China}
}

\renewcommand{\shortauthors}{Zhan, et al.}

\begin{abstract}
Text-to-image generation systems have emerged as revolutionary tools in the realm of artistic creation, offering unprecedented ease in transforming textual prompts into visual art. However, the efficacy of these systems is intricately linked to the quality of user-provided prompts, which often poses a challenge to users unfamiliar with prompt crafting. This paper addresses this challenge by leveraging user reformulation data from interaction logs to develop an automatic prompt reformulation model. Our in-depth analysis of these logs reveals that user prompt reformulation is heavily dependent on the individual user's capability, resulting in significant variance in the quality of reformulation pairs. To effectively use this data for training, we introduce the Capability-aware Prompt Reformulation (CAPR) framework. CAPR innovatively integrates user capability into the reformulation process through two key components: the Conditional Reformulation Model (CRM) and Configurable Capability Features (CCF). CRM reformulates prompts according to a specified user capability, as represented by CCF. The CCF, in turn, offers the flexibility to tune and guide the CRM's behavior. This enables CAPR to effectively learn diverse reformulation strategies across various user capacities and to simulate high-capability user reformulation during inference. Extensive experiments on standard text-to-image generation benchmarks showcase CAPR's superior performance over existing baselines and its remarkable robustness on unseen systems. Furthermore, comprehensive analyses validate the effectiveness of different components. CAPR can facilitate user-friendly interaction with text-to-image systems and make advanced artistic creation more achievable for a broader range of users.
\end{abstract}




\maketitle


\section{Introduction}

In the realm of intelligent information systems, effective communication between users and systems is important. Traditionally, this interaction has been facilitated through queries in search engines, serving as concise yet powerful instructions to retrieve relevant information~\cite{silverstein1999analysis, chau2005analysis}. With the advent of Artificial Intelligence Generated Content (AIGC) systems like Midjourney~\cite{midjourney2023}, these instructions have evolved into prompts, a critical element in shaping the quality and relevance of system responses~\cite{parsons2022dalle, oppenlaender2022taxonomy}. Despite their importance, most users struggle to craft optimal queries or prompts~\cite{witteveen2022investigating, xie2023prompt}, making automatic reformulation techniques an essential component for enhancing system performance~\cite{openai2023improving, hao2022optimizing, brade2023promptify}. Similarly, in the domain of search engines, techniques such as query auto-completion~\cite{cai2016survey}, expansion~\cite{carpineto2012survey}, and suggestion~\cite{sordoni2015hierarchical} have significantly improved user experience and system efficacy, becoming indispensable features of commercial search engines~\cite{boldi2011query}.

While the benefits of query reformulation are well-established within search engines~\cite{baeza2004query, jiang2014learning, dehghani2017learning, boldi2011query}, the exploration of prompt reformulation for AIGC systems, particularly text-to-image generation systems, is relatively limited. Text-to-image generation systems have revolutionized the field of artistic creation, simplifying the process to unprecedented ease~\cite{kingma2021variational, ho2020denoising, lu2023specialist, liao2022text, kang2023variational, zhou2023shifted, kang2023scaling}. They operate by converting user-provided text prompts into visual imagery. However, their efficacy heavily depends on the quality of the input prompts. Effective prompts should conform to a specific format, precisely describes the scene, and consist of professional terminologies such as artist names~\cite{brade2023promptify, witteveen2022investigating, parsons2022dalle}. This level of complexity in prompt writing is challenging, making learning and practice necessary for users~\cite{liu2022design, deckers2023manipulating}. Users often rely on studying exemplary prompts shared within the community~\cite{stablediffusion_prompts, prompthero}, consulting guides on effective prompting~\cite{parsons2022dalle, oppenlaender2022taxonomy}, and engaging in trial-and-error to refine their skills~\cite{xie2023prompt}. This high learning cost and constant need to reformulate prompts substantially affect the user experience.
 
In this paper, we focus on leveraging user-generated reformulation data from interaction logs to develop an automatic prompt reformulation model. This is motivated by the observation that users dedicate considerable effort to prompt reformulation~\cite{xie2023prompt}, creating rich data in the interaction logs. By designing a model that builds upon these user efforts, we aim to significantly reduce the burden of manually reformulating prompts and substantially improve the user experience.

Our initial analysis of the user interaction logs reveals a significant distinction between query reformulation and prompt reformulation in text-to-image generation scenarios. Unlike query reformulation, where users benefit significantly from search results to reformulate their queries~\cite{chen2021towards, chen2021hybrid, guan2013utilizing}, the effectiveness of prompt reformulation for text-to-image systems relies heavily on the individual user's capability, rather than feedback from the system. Such user capability, which varies widely among individuals and generally remains consistent within a single session, leads to a wide spectrum of prompt quality and predominantly marginal reformulation improvements. For example, the initial prompts of some users may substantially surpass the reformulated prompts of others, and instances of poorly crafted initial prompts being significantly improved through reformulation are remarkably rare. This scenario contrasts sharply with query reformulation scenarios, where users often successfully find the relevant information by the end of a session~\cite{dehghani2017learning, chen2021towards}. Consequently, unlike previous query reformulation studies that often disregard user capability in model design, we aim to design a novel framework to introduce the crucial influence of user capability in the process of prompt reformulation.

To address this challenge, we propose the Capability-aware Prompt Reformulation framework (CAPR). CAPR incorporates user capability into the reformulation process, thereby enabling effective training with user-generated data from interaction logs. It consists of two foundational components: the Conditional Reformulation Model (CRM) and Configurable Capability Features (CCF). CRM is adept at tailoring prompt reformulation according to a specified user capability, as represented by CCF. The CCF, in turn, offers the flexibility to tune and guide the CRM's behavior. The two components offer two key benefits: (1) CRM, by adapting to user capability, aligns closely with the nature of interaction log data and thus can harvest a wealth of prompt reformulation skills across different user capability levels. (2) CCF, by introducing scrutable generation capability features, allows us to control the quality of inference and generate high-quality prompts. Consequently, CRM is empowered to surpass the average user capabilities in the training data, thereby yielding superior reformulation outcomes in practical applications\footnote{Code is open-sourced at \url{https://github.com/jingtaozhan/PromptReformulate}}.

We conduct comprehensive experiments on standard text-to-image generation benchmarks to examine the efficacy of CAPR. Results suggest that CAPR substantially outperforms a variety of baselines, including generic language models like GPT4 and various reformulation models. Its effectiveness can also generalize to an unseen text-to-image generation system, demonstrating its robustness. A detailed ablation study also shows that CPR can generate target images based on the specified capability conditions.

In summary, our contributions are in three folds:
\begin{itemize}
	\item To the best of our knowledge, this is the first study that utilizes interaction logs to train a prompt reformulation model for text-to-image generation.
	\item We provide a comprehensive analysis that differentiates prompt reformulation in text-to-image generation from traditional query reformulation in search engines
	\item  Inspired by our analysis, we propose a novel prompt reformulation model tailored for training on prompt reformulation logs. Results demonstrate that it achieves state-of-the-art reformulation performance on standard benchmarks.
\end{itemize}

\section{Related Work}

\subsection{Text-to-Image Generation}
Text-to-image generation is a rapidly evolving field in artificial intelligence that focuses on creating visual images from textual descriptions. This technology has gained considerable attention, particularly in the realm of digital art creation, as exemplified by systems like Midjourney~\cite{midjourney2023} and DALLE~\cite{openai2023improving}. Existing text-to-image systems mostly adopt Diffusion as the model architecture~\cite{ho2020denoising, kingma2021variational}, which can progressively transform a random noise into a coherent image with a text as guidance. The training of these diffusion models relies heavily on extensive datasets comprising images coupled with descriptive captions~\cite{schuhmann2022laion}. During training, the model learns to correlate textual descriptions with visual features, enabling it to generate relevant images for a given text input. A noteworthy aspect of this training process is that high-quality web images are usually accompanied by professional-level captions consisting of artist names and photography terminologies. Consequently, the trained models tend to favor such prompts~\cite{ghosh2023mttn, witteveen2022investigating}, which, unfortunately, are difficult to write for average users.

\subsection{Query Reformulation}
Query reformulation stands as an important technique in enhancing user interaction with search engines. It addresses common issues where initial queries fail to search relevant results~\cite{boldi2011query}. Techniques like auto-completion and query suggestion play a crucial role in assisting users to refine their queries, thereby enhancing the likelihood of retrieving relevant information. Prior query reformulation analyses have shown that feedback from search systems helps provide relevant terms and plays a significant role in aiding query reformulation~\cite{chen2021towards, chen2021hybrid}. Based on this, many query reformulation methods leverage terms from search results to modify the initial queries~\cite{salton1990improving, guan2013utilizing}. Moreover, some research focuses on developing reformulation models based on query logs~\cite{dehghani2017learning, chen2021towards}. This approach typically assumes that users successfully find relevant information by the end of their session. Thus, it constructs training pairs by treating the final query in a session as a target label for training. However, as discussed in Section~\ref{sec:analysis_prompt_reform}, the assumptions about the system's feedback and user reformulation data in general web search scenarios do not hold for text-to-image generation. Therefore, a tailored methodology for prompt reformulation is necessary.

\subsection{Text-to-Image Prompts}
Prompt quality is a crucial factor in the effectiveness of text-to-image generation systems~\cite{openai2023improving, hao2022optimizing, brade2023promptify}. Crafting a high-quality prompt, however, poses a significant challenge. It not only demands rich art knowledge like artist names and style elements but also typically requires a time-consuming iterative process of tuning and refinement~\cite{brade2023promptify, witteveen2022investigating, parsons2022dalle}. To aid users in this endeavor, various online platforms have emerged, offering spaces for sharing well-crafted prompts~\cite{stablediffusion_prompts, prompthero}. In addition, comprehensive guides and textbooks have been written to teach prompt crafting techniques~\cite{parsons2022dalle, oppenlaender2022taxonomy}. There are even marketplaces dedicated to trading high-quality prompts~\cite{shen2023prompt}. However,  these resources often come with substantial demands of either time investment or financial cost. To alleviate the complexity of prompt crafting, it is important to develop an automated model that is capable of reformulating subpar prompts into well-crafted ones. The primary obstacle in developing such a model is the difficulty in annotating prompt reformulation pairs for training, which requires annotators with extensive experience, leading to high costs and complexities in labeling. Previous researchers bypass this obstacle by constructing synthetic refinement data~\cite{hao2022optimizing}. They crawl high-quality prompt demonstrations from the internet and rephrase them to user languages. However, the quality of such synthesized data is subpar, which harms the performance of the trained model. Considering this issue, our approach aims to extract reformulation data from readily available interaction logs, offering a more direct and practical solution.

\section{Problem Formulation}

This section formulates the core challenge for prompt reformulation in text-to-image generation scenarios. 

Text-to-Image generation systems~\cite{kingma2021variational, ho2020denoising, kang2023variational}, such as Midjourney and DALL-E, represent cutting-edge intelligent tools for artistic creation. These systems work by transforming textual descriptions, known as prompts, into visual imagery~\cite{lu2023specialist, liao2022text, zhou2023shifted, kang2023scaling}. At each interaction round, the user provides a prompt that describes their envisioned image. The system interprets this text to generate a corresponding image. While these systems liberate users from the technicalities of traditional art creation, they demand high-quality prompts to accurately generate users' envisioned images. Mathematically, let $p$ be the prompt and $i$ be the rendered image. The text-to-image generation system is denoted as $\mathcal{G}$, with $\mathcal{G}(i|p)$ signifying the probability of generating image $i$ from prompt $p$. 

Evaluating the generation quality has been extensively studied. The ideal approach involves human annotators, but without a large team and comprehensive guidelines, human's diverse preferences can lead to inconsistent assessments~\cite{wu2023humanv2, chen2018image}. To avoid this, prior research has developed automatic evaluation models  to simulate human preferences by training them on extensive human annotations~\cite{aesthetic_predictor, xu2023imagereward, wu2023humanv2}. In this study, we employ these automatic scoring models for evaluation. We denote the scoring model as $f$, where $f(p, i)$ indicates the likelihood of user satisfaction with the generated image $i$ for prompt $p$. Consequently, the generation quality for a prompt $p$ can be quantified as $\E [ \mathcal{G}(i|p) \cdot f(p, i) ]$. 

Text-to-image generation systems are usually sensitive to input prompts, making prompt crafting a form of ``art''~\cite{brade2023promptify, witteveen2022investigating, parsons2022dalle}, a skill that many users do not possess.
Automatic prompt reformulation stands as a critical solution to this challenge. A reformulation model acts as an intermediary, transforming an initial user prompt into a version better suited for the text-to-image system. For example, the reformulation model can refine vague descriptions and enrich the prompt with artist references. We use $\Omega$ to represent a reformulation model, and $\Omega(\hat{p}|p)$ is the probability of reformulating $p$ to $\hat{p}$. The reformulation model is expected to maximize the generation quality, which is formulated as:
\begin{equation}
\label{eq:evaluation}
	\max_{\Omega} \E [ \Omega(\hat{p}|p) \cdot \mathcal{G}(\hat{i}|\hat{p}) \cdot f(p, \hat{i}) ]
\end{equation}
The reformulated prompt $\hat{p}$ serves as an intermediate in the generation process. The evaluation scores are computed based on the initial prompt and the images. The reformulated prompt aims to help render better images.

\begin{table}[t]
\centering
\small
\begin{tabular}{l|l|l}
\toprule
\textbf{Notion} & \textbf{Text-to-Image Generation} & \textbf{Search Scenario} \\ \midrule
\( p \) & Prompt & Query \\ 
\( i \) & Rendered image & Search result page \\ 
\( \mathcal{G} \) & Text-to-image system & Search engine \\ 
\( f \) & Satisfaction with image & Relevance of search results \\ 
\( \Omega \) & Prompt reformulation model & Query reformulation model \\ 
\bottomrule
\end{tabular}
\caption{Summary of the notions for text-to-image generation and Search Engine Scenarios.}
\label{tab:notions}
\end{table}

To clarify the introduced notions, we summarize them in Table~\ref{tab:notions}. The table also shows the corresponding interpretations in search engine scenarios, which will be elaborated in the next section where we compare prompt reformulation with query reformulation.

\section{Analysis of Prompt Reformulation}
\label{sec:analysis_prompt_reform}
This section deeply analyzes how users reformulate prompts, which serves as a crucial insight for training a reformulation model on interaction logs. We first compare prompt reformulation with query reformulation. Analysis results reveal that prompt reformulation is influenced by user capability to a larger extent. Then, we validate this observation through an examination of reformulation behaviors in large-scale interaction logs.

\subsection{Reformulation: Prompt vs. Query}

Prompt and query reformulation share structural similarities, which facilitates examining prompt reformulation through the lens of established findings in query reformulation. As depicted in Table~\ref{tab:notions}, the components in prompt and query reformulation can correspond to each other. For instance, the rendered image corresponds to a search result page, and the generation model parallels a search engine. In both contexts, users describe an information need through textual input, guiding the system to produce the desired output.

In query reformulation, a well-established body of research identifies three pivotal factors that facilitate the process~\cite{chen2021towards, chen2021hybrid, dehghani2017learning, guan2013utilizing}: the initial query, the user's understanding towards search engine mechanism, and the search result page. These factors are foundational in guiding effective query reformulation. The initial query lays the groundwork, providing a basis for subsequent refinements. Knowledge of the search engine's mechanism enables users to understand how their queries are processed and related to the results. This knowledge helps them to effectively optimize the queries further. Moreover, the search results themselves often provide precise terminology and relevant context, assisting users in refining their queries with greater specificity~\cite{efthimiadis2000interactive, spink2000use}.

\begin{figure}
  \centering
  \includegraphics[width=0.999\linewidth]{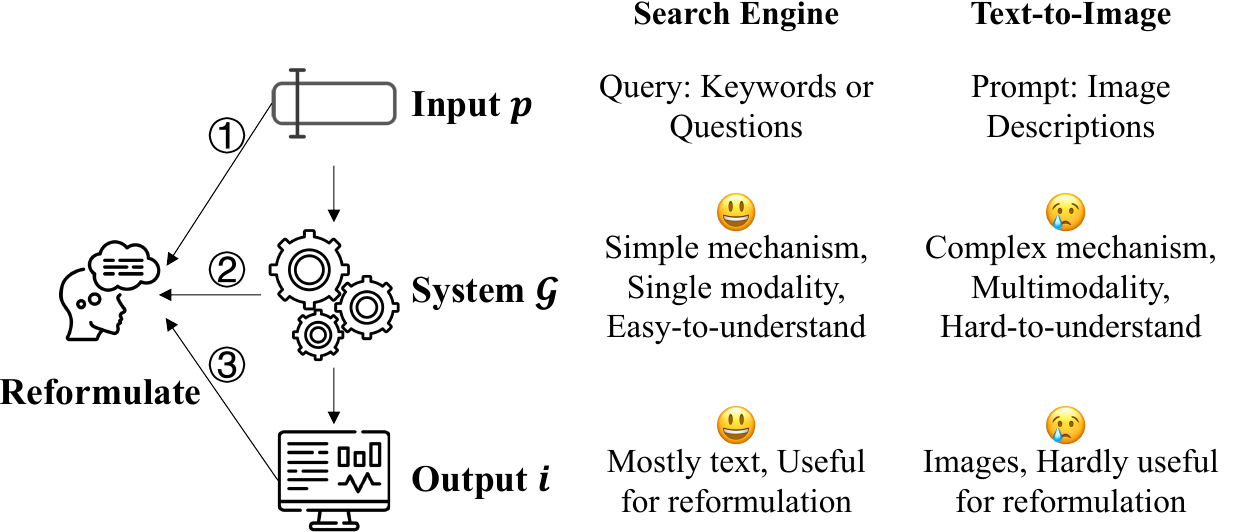}
  \caption{
  	Comparing prompt reformulation with query reformulation in terms of three key factors: \ding{172} the initial input \ding{173} user's understanding of the system's mechanics \ding{174} the previous system's output. The latter two can hardly help users reformulate better prompts, indicating that prompt reformulation is a more challenging task for users.
  	\label{fig:reform_query_vs_prompt}
  }
\end{figure}

Conversely, in the realm of text-to-image generation, the last two facilitative factors are absent, making prompt reformulation a substantially more challenging task. We compare prompt reformulation with query reformulation in Figure~\ref{fig:reform_query_vs_prompt}. Unlike search engines, whose operating mechanisms (keyword matching, single modality) are obvious and generally understood by users, text-to-image systems often operate as ``magical black boxes''. The intricate neural computations and the multimodality nature are typically beyond the user's comprehension. Furthermore, the output of these systems, being visual imagery, does not offer textual cues, e.g., artist names or stylistic terminologies, that users can directly add to their prompts. The lack of informative feedback from the output, combined with the black-box nature of the generation systems, places a significant burden on users. They have to rely heavily on their inherent capability to intuit and imagine how different textual inputs might influence the visual output. This is a process that is less guided and more speculative compared to the more systematic and feedback-oriented process of query reformulation.

\subsection{Investigation of Prompt Session Log}

\begin{figure}[t]
    \centering
    \subfloat[Overall Generation Quality measured by ImageReward~\cite{xu2023imagereward}]{
        \includegraphics[width=.45\columnwidth]{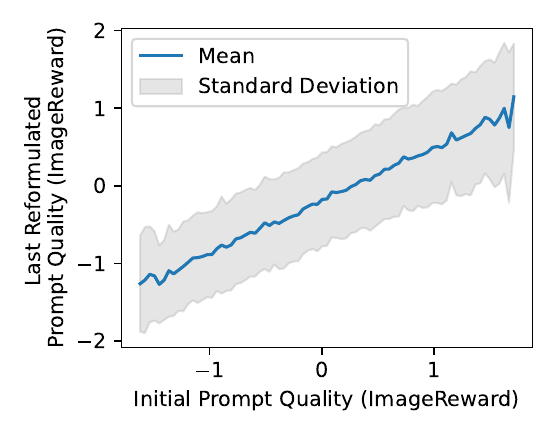}
        \label{fig:session_imaegreward}
    }
    \hfill
    \subfloat[Aesthetic Generation Quality measured by Aesthetic Predictor]{
        \includegraphics[width=.45\columnwidth]{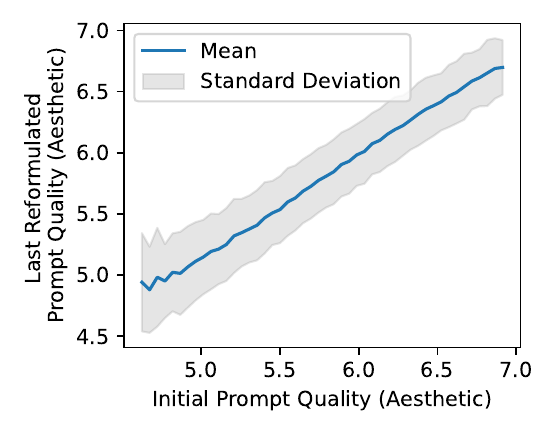}
        \label{fig:session_aesthetic}
    }
    \caption{Comparison of generation quality between initial and reformulated prompts within a session, evaluated by ImageReward~\cite{xu2023imagereward} and Aesthetic scoring models~\cite{aesthetic_predictor}. Results reveal limited quality improvement through users' reformulation, suggesting that prompt quality largely depends on the user's initial capabilities. Session contexts such as generation feedback usually offer limited assistance.}
    \label{fig:session_analysis}
\end{figure}

To validate the above analysis, we investigate user reformulation behaviour from an interaction log. The results demonstrate that prompt reformulation is indeed a challenging task that heavily depends on users' inherent capabilities. We first introduce the dataset, our analysis methodology, and then elaborate on our findings. 

\subsubsection{Dataset} 
\label{sec:investigate_log_dataset_info}
In this analysis, we utilize DiffusionDB~\cite{wang2022diffusiondb}, a comprehensive log capturing 1.8 million interactions from 10 thousand users. This dataset includes prompts submitted by users, images generated by the system, user IDs, and timestamps. Its extensive scale and diversity enable a robust analysis of real user reformulation behavior. Since the dataset does not split interactions into sessions, we construct sessions based on timestamps and prompt topics. Specifically, inspired by the construction of search sessions, adjacent prompts that are submitted by the same user within 20 minutes and surpass a similarity threshold of 0.1 (as determined by the CLIP model~\cite{radford2021learning}) are classified into the same session. We have manually examined several session splits constructed through this way. The quality of most sampled sessions is reasonable and reliable. Eventually, we obtain 30k sessions in total.

\subsubsection{Analysis Methodology}
We use the first prompt of each session to represent the user's original intent and investigate how the last prompt of a session improves the generation quality. We use scoring models, namely ImageReward~\cite{xu2023imagereward} and Aesthetic Predictor~\cite{aesthetic_predictor}, to assess the quality of images generated from both the initial and the last prompts. ImageReward considers relevance and aesthetic quality, reflecting overall user satisfaction, while Aesthetic Predictor focuses solely on the visual appeal of the generated images. The results are depicted in Figures~\ref{fig:session_imaegreward} and \ref{fig:session_aesthetic}.

\subsubsection{Empirical Findings}
The empirical results, as illustrated in Figure~\ref{fig:session_analysis}, show that the quality of user reformulation exhibits significant variance and that most users face challenges in substantially improving their initial prompts in the sessions. This is primarily attributed to the factors discussed earlier: the limited assistance from system feedback in prompt reformulation and the substantial dependence on the user's inherent prompt-writing capability. Notably, this capability tends to remain static within a single session, leading to a scarcity of cases where users successfully reformulate poorly crafted initial prompts to high-quality ones. 

Given this observed variability in reformulation quality and the prevalence of suboptimal reformulation pairs, the conventional approach of training reformulation models based on such user data presents significant challenges. Traditional query reformulation studies typically employ a direct sequence-to-sequence translation model based on reformulation pairs~\cite{dehghani2017learning, halder2020modeling, chen2020exploring}, but such a method may not work well in our context. Training a model on these inconsistent and suboptimal pairs could lead to substantial confusion and eventually diminish the model's overall effectiveness. If we filter the dataset to include only optimal reformulation pairs, the final size of the training data would be too small to ensure robust model training. Thus, new methods are needed to address these challenges in text-to-image prompt reformulation.

\section{Methodology}

Inspired by the insights gained from our previous analysis, we propose the Capability-aware Prompt Reformulation (CAPR) framework, a novel approach to effectively train a prompt reformulation model using human-generated reformulation data. CAPR innovatively incorporates a condition that mirrors user capability into the process of prompt reformulation. This design aligns seamlessly with our findings regarding the dependency of user reformulation on their capabilities. This unique approach ensures that CAPR is not excessively influenced by the inconsistent reformulation qualities prevalent in the training data. Instead, it enables the framework to adeptly learn diverse reformulation techniques from users with different levels of expertise. Furthermore, the condition representing user capability is adjustable, providing CAPR with the flexibility to function at a high level during inference. This allows CAPR to deliver high-quality prompt reformulations that surpass the average levels in its training dataset.

\subsection{Model Architecture}

\begin{figure}
  \centering
  \includegraphics[width=0.9\linewidth]{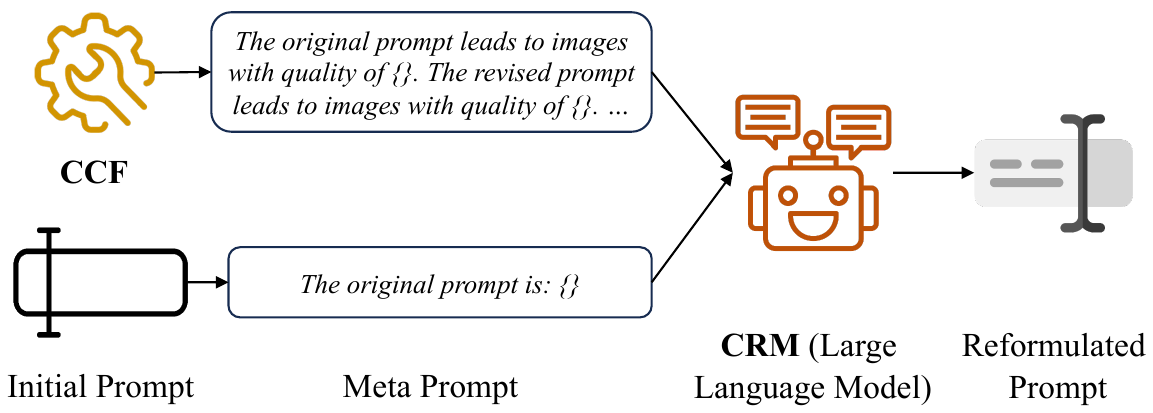}
  \caption{
  Architecture of Capability-aware Prompt Reformulation (CAPR). It consists of two components: the Conditional Reformulation Model (CRM) and Configurable Capability Features (CCF). Given a certain user capability indicated by CCF, CRM reformulates prompts accordingly.
  \label{fig:overall_cpr}
  }
\end{figure}

In this section, we describe the overall architecture of CAPR.
CAPR introduces the user capability into the reformulation process by decomposing a reformulation model into two components: a Conditional Reformulation Model (CRM) and Configurable Capability Features (CCF). The model architecture is illustrated in Figure~\ref{fig:overall_cpr}. CCF is designed to represent various levels of user capacities in prompt writing, while CRM specializes in reformulating prompts in accordance with the specified capability. Mathematically, we denote CRM by $\omega$ and CCF by $c$. The reformulation process $\Omega$ is thus decomposed as follows: 
\begin{equation}
\label{eq:cpr_decompose}
	\Omega(\hat{p} | p) = \sum_c \omega(\hat{p} | p, c) \cdot {\rm P}(c)
\end{equation}
Here, $p$ represents the original prompt, and $\hat{p}$ denotes the reformulated prompt. During training, by introducing user capability $c$ as an additional condition, CRM ($\omega$) can adapt to the various qualities of reformulation pairs and effectively learn different reformulation strategies. During inference, by configuring the distribution of CCF (${\rm P}(c)$), CCF can enable CRM to simulate a high-capability user to reformulate prompts, transcending the limitations of the training data and generating reformulations with high expertise. Next, we describe the specific model designs.

\subsubsection{Conditional Reformulation Model (CRM)}
CRM should effectively interpret the input user capability condition and reformulate the prompt accordingly. For this purpose, we implement it as a large language model due to its remarkable conditional generation abilities. As illustrated in Figure~\ref{fig:overall_cpr}, the initial prompt and the user capability condition are transformed into textual formats, termed ``meta prompts''. These meta prompts are then concatenated and fed into a large language model to ensure that the task's nature and input details are clearly described, enabling the language model to accurately comprehend the conditions and produce conditional outputs. 

\subsubsection{Configurable Capability Features (CCF)}
\label{sec:ccf_model}
CCF reflects the user's capability to effectively reformulate prompts. Given that such capacities are not explicitly recorded in interaction logs, CCF should be designed to be computable from reformulation pairs. Besides, it should also be easily tunable to guide the CRM toward simulating high-quality reformulations during inference. For these purposes, we employ a suite of scoring models to evaluate the quality of both the initial and reformulated prompts and use the output scores as CCF. This approach not only offers a direct assessment of the reformulation capability but also facilitates the straightforward extraction of user capacities from the interaction logs. The clarity and quantifiable nature of these metrics ensure they can be easily tuned during inference to lead CRM toward high-level reformulations. Specifically, CCF encompasses the following features:
\begin{itemize}
\item \textbf{Overall quality}: It measures the likelihood of user satisfaction with images generated from their prompts, serving as an indicator of overall prompt-reformulation ability. ImageReward model~\cite{xu2023imagereward} is used to predict user satisfaction levels based on the generated results.
\item \textbf{Prompt-image similarity}: It assesses the ability of a user’s prompt to yield coherent images, a key aspect of generation quality. The CLIP model~\cite{radford2021learning} is employed to assess the coherence between the prompt and the rendered image.
\item \textbf{Aesthetic quality}: It indicates the visual appeal of the images produced from the prompts, a key aspect of generation quality. An aesthetic predictor~\cite{aesthetic_predictor} is used to evaluate the visual appeal of the generated images. 
\item \textbf{Prompt Length}: It reflects the user's skill in creating detailed prompts, a key aspect of prompt-writing capability. It is measured by the number of comma-separated phrases.
\end{itemize}

\subsubsection{Integrating CRM and CCF}
To effectively integrate CRM with CCF, we construct a ``meta prompt'' using a structured template. This template is designed to contextualize the numeric features, making them more interpretable for the language model. The template used in our experiments is: 

\textit{``A text-to-image generation system transforms text prompts into visual images. The effectiveness of this conversion depends on the prompt. The original prompt leads to images with prompt-image similarity of \{\}, aesthetic quality of \{\}, and overall quality of \{\}. To improve these metrics, new images are generated based on a revised prompt. After evaluating the new images for the initial prompt, the updated scores are: prompt-image similarity of \{\}, aesthetic quality of \{\}, and overall quality of \{\}. The revised prompt is structured into \{\} phrases, each separated by a comma. Considering the given information, the revised prompt should be:''}

Next, we present the details of training the CAPR model.

\subsection{Learning Conditional Generation}

\begin{figure}
  \centering
  \includegraphics[width=0.999\linewidth]{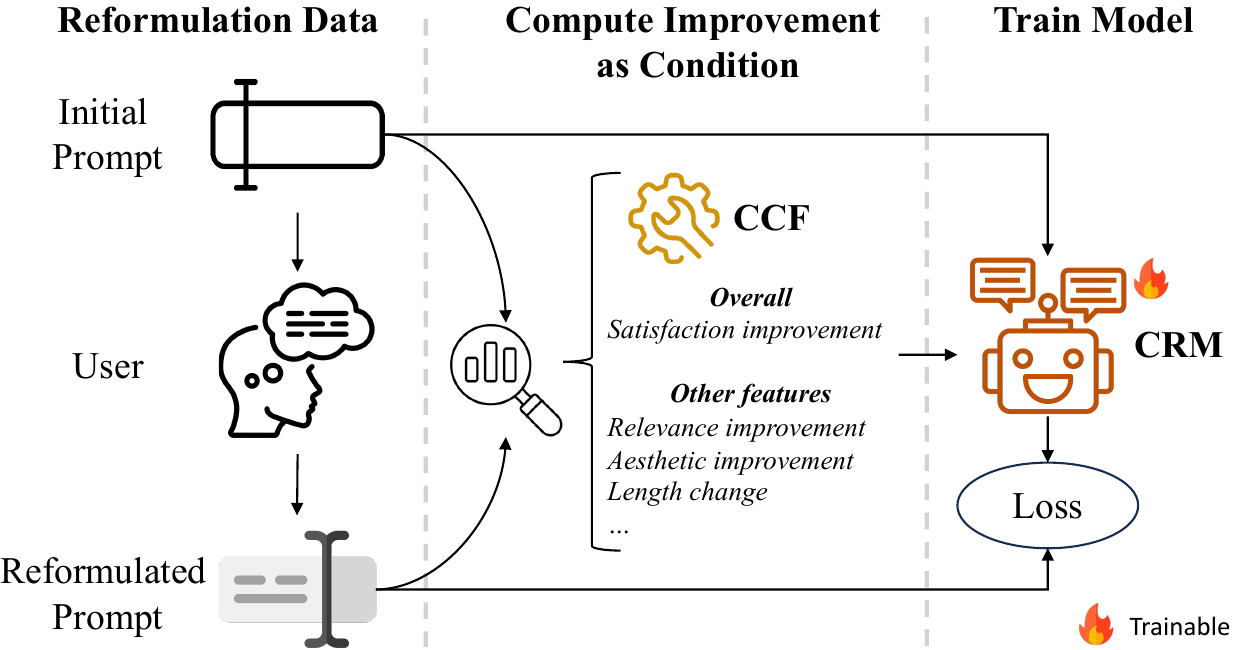}
  \caption{
  Training process of Capability-aware Prompt Reformulation (CAPR). Configurable Capability Features (CCF) is computed based on the training pairs, and Conditional Reformulation Model (CRM) is trained to predict the reformulated prompt given the initial prompt and CCF.
  \label{fig:train_cpr}
  }
\end{figure}

In this subsection, we detail the training process of the Conditional Reformulation Model (CRM). As illustrated in Figure~\ref{fig:train_cpr}, CRM training has two distinct stages, which are introduced below.

\subsubsection{Computing Configurable Capability Features (CCF)} \mbox{}

CRM's training requires the creation of data triplets: an initial prompt ($p$), its reformulation ($\hat{p}$), and a corresponding capability condition ($c$). While $p$ and $\hat{p}$ are directly sourced from user inputs within the interaction logs, $c$ is derived from each reformulation pair. As outlined in the previous subsection, we employ a suite of scoring models to assess the quality of both the initial and reformulated prompts, utilizing these evaluations as the basis for CCF.

The derived scores for overall quality, prompt-image similarity, and aesthetic quality are initially in floating-point formats with diverse ranges. To facilitate the language model's interpretation of these metrics, we convert them into a uniform integer scale. This is achieved by quantizing the scores into $K$ integer values, ranging from $0$ to $K-1$. The quantization is executed by evenly distributing the range of scores from the minimum to the maximum into $K$ discrete intervals. In our experiments, we find CRM is robust to different $K$ and empirically set $K$ to 10.

\subsubsection{Training Conditional Reformulation Model (CRM)}
The training of the CAPR model is guided by an autoregressive language modeling loss function. For each triplet of $p$, $\hat{p}$, and $c$, the model is trained to predict $\hat{p}$ given $p$ and $c$. Mathematically, it is trained to minimize the following loss:
\begin{equation}
	\mathcal{L} = - {\rm log} \; \omega(\hat{p} | p, c) = - \sum_n {\rm log} \; \omega(\hat{p}_n | \hat{p}_{1:n-1}, p, c)
\end{equation}
where $\hat{p}_n$ is the $n$-th token of prompt $\hat{p}$. This process makes CAPR to learn different kinds of reformulation skills and accurately reformulate prompts according to the specified condition.

\subsection{Configuring Capability Features}

\begin{figure}
  \centering
  \includegraphics[width=0.999\linewidth]{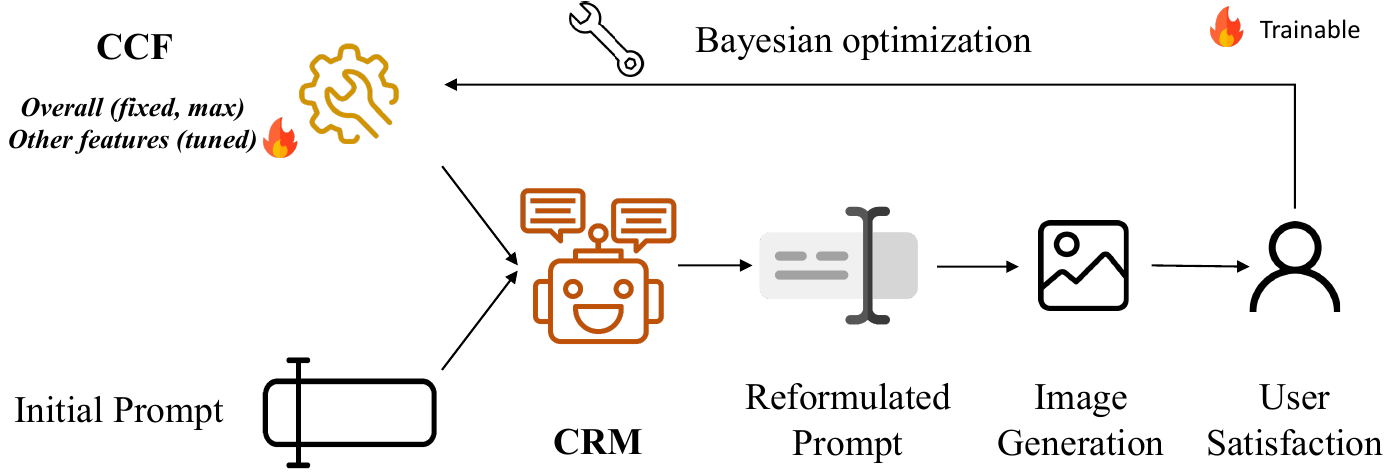}
  \caption{
  Configuration of Configurable Capability Features (CCF). The Conditional Reformulation Model (CRM) has been trained and is frozen. Within CCF, the overall quality metric is set to the highest, and other features of CCF are tuned to maximize the generation quality. The tuning process is accelerated with Bayesian optimization.
  \label{fig:infer_cpr}
  }
\end{figure}

After completing the training for CRM, we focus on optimizing the Configurable Capability Features (CCF) to enhance reformulation performance. As introduced in Section~\ref{sec:ccf_model}, CCF for each reformulation pair incorporates two dimensions: quality assessments of the initial prompt and the reformulated prompt, referred to as $c'$ and $c''$, respectively. These two components together form the composite CCF, denoted as:
\begin{equation}
	c = (c', c'')
\end{equation} 
Next, we describe how to configure them, respectively.

\subsubsection{Efficient Prediction of Initial Prompt Quality}
The assessment of the initial prompt's quality $c'$ is executed efficiently within our framework. While the ideal method would involve generating and evaluating images based on the initial prompt, such a procedure is not practically feasible during inference because of the time-consuming generation process. To address this, we train a RoBERTa-Large model to predict the prompt quality based simply on the prompt. The trained model directly predicts the prompt's overall quality, its similarity to generated images, and aesthetic appeal during inference. While this deviates slightly from the precise quality scores, it significantly accelerates the evaluation process by saving the generation time. In practice, it reduces the time required from about 10 seconds to a mere 10 milliseconds.

\subsubsection{Optimizing Expected Reformulation Quality} \mbox{}
\label{sec:tune_expected_ccf}

The other aspect of CCF is the expected quality for reformulated results $c''$. During the training phase, this aspect of CCF serves as a guide for CRM by indicating the desired quality level of the generated prompt. During inference, this feature should be carefully set so that CRM is guided to perform optimally.

In an ideal scenario, setting this feature to its maximum guides the CRM to produce the most optimal prompt possible. However, this is not practical in practice due to the limits of the training data. Since few reformulated prompts in the training data can simultaneously achieve the max scores in terms of all CCF features (overall quality, similarity, and aesthetic quality), setting $c''$ to the maximum value pushes CRM to an extremely out-of-distribution scenario and potentially compromises its performance. 

Therefore, we search for the best $c''$ that can maximize user satisfaction. This is depicted in Figure~\ref{fig:infer_cpr} and can be mathematically formulated as follows:
\begin{equation}
\begin{aligned}
 	c''^* & = \arg \max_c \phi (c'') \\
 	\phi(c'') &=  \E_{p, \hat{p}, i} [ \omega(\hat{p} | p, (c', c'')) \cdot \mathcal{G}(i | \hat{p}) \cdot f(p, i) ]
\end{aligned}
\end{equation}
Here, $\mathcal{G}$ represents the generation system, and $f$ measures user satisfaction, with $\phi(c'')$ indicating the average performance for a specified condition $c''$. While it would be ideal to tailor $c''$ for each individual prompt, it is impractical due to time and resource constraints. To simplify this process, we adopt several strategies:
\begin{itemize}
	\item Search beforehand: By constructing a validation set, we determine the best configuration for $c''$ prior to the inference stage and thus eliminate the need for repetitive searches during individual prompt evaluations.
	\item Adaptive Reparameterization: To accommodate the diverse nature of prompts while maintaining pre-inference searching efficiency, we reparameterize $c''$ as:
	\begin{equation}
		\label{eq:ccf_reparameterization}
		c'' = c' + \delta 
	\end{equation}
	$\delta$ represents the expected improvement for CRM. We search for the optimal $\delta$ instead of $c''$. This allows tailoring $c''$ to individual prompt qualities without a one-size-fits-all $c''$.
	\item Search Space Reduction: Given the meaningful nature of each feature within CCF, we can heuristically narrow the search space. With our primary goal being overall user satisfaction, we set the expected overall quality factor to its maximum while searching for other features.
	\item Advanced Search Techniques: We employ Bayesian optimization for its efficiency and effectiveness. This method models $\phi(c'')$ as a Gaussian process, dynamically exploring new condition values based on previous results to refine the search progressively. Interested readers can refer to the skopt toolkit~\cite{head_2021_5565057} for more details.
\end{itemize}
Through these optimization strategies, we can narrow down the search space from around $10^4$ to $10^3$ and use Bayesian optimization to conduct an effective search within 50 calls. 

It is also important to note that the determined $c''$ is optimized for general user satisfaction. However, users with specific requirements, such as a focus on aesthetic quality, can use the provided $c''$ as a starting point and further customize the capability values to suit their needs. The flexibility and interpretability of CCF allow for easy user-driven adjustments in prompt reformulation behavior.

\section{Experiments}

\subsection{Experimental Setup}

\subsubsection{Training Data}

We use a large-scale interaction log named DiffusionDB~\cite{wang2022diffusiondb} for training. The dataset covers the real interactions from the official Stable Diffusion Discord channel for half a month. In total, it logs 1.8 million interactions from 10k users. When the interaction log was constructed, the text-to-image generation system used Stable Diffusion 1.4 model~\cite{rombach2022high} for generation. We split the interactions into sessions, which is detailed in Section~\ref{sec:investigate_log_dataset_info}. We use the initial and the last prompts of a session to construct reformulation pairs. In total, we construct 30k reformulation pairs.  

\subsubsection{Evaluation Setup}

We conduct evaluations using the HPSv2 benchmark dataset~\cite{wu2023humanv2}.
The dataset includes a wide variety of prompts categorized into Anime, ConceptArt, and Painting. Each category contains 800 prompts. For each prompt, we generate four images to ensure a robust assessment.
We employ automated scoring models for assessments, including ImageReward~\cite{xu2023imagereward} and HPSv2~\cite{wu2023humanv2}. They are trained to mimic human preferences and have been demonstrated to be reliable in evaluating text-to-image generation quality. 
We use two text-to-image generation models to evaluate the reformulation effectiveness, including Stable Diffusion 1.4 (SD1.4)~\cite{rombach2022high} and Stable Diffusion XL base 1.0~(SDXL)~\cite{podell2023sdxl}. SD1.4 is exactly the system used when the interaction log was constructed and therefore is a seen system to our model. SDXL is a recently released state-of-the-art generation system and thus is an unseen system. 

\newcommand{\B}{\bfseries}
\newcommand{\allsig}{$^{*\dagger}$}
\newcommand{\sigbase}{$^{*}$\;\,}
\newcommand{\sigfilter}{$^{\dagger}$\;}
\newcommand{\nosig}{\;\;\,\,}

\begin{table*}
\centering
\small
\caption{Reformulation performance on the seen system~(SD1.4). ImageReward~\cite{xu2020neural} and HPSv2~\cite{wu2023humanv2} serve as evaluation models ($f$ in Eq.~(\ref{eq:evaluation})) and numbers are the average output scores.
$^{*}$ and $^{\dagger}$ separately indicate that performance is significantly better than SD1.4 and SD1.4+PR-Filter at $p < 0.01$ level measured by ttest. CAPR significantly outperforms baselines.}
\label{tab:compare_on_sd14}
\setlength{\tabcolsep}{3mm}
\begin{tabular}{l|ccc|ccc}
   	\toprule
    \multirow{2}{*}{Method} & \multicolumn{3}{c|}{ImageReward} & \multicolumn{3}{c}{HPSv2} \\
    & Anime & ConceptArt & Painting & Anime & ConceptArt & Painting \\
    \midrule
SD1.4~\cite{rombach2022high}                          & 0.038\nosig & 0.185\nosig & 0.190\nosig & 27.42\nosig & 26.86\nosig & 26.86\nosig \\
+ GPT3.5~\cite{brown2020language}                       & -0.037\nosig & 0.030\nosig & 0.126\nosig & 27.36\nosig & 26.77\nosig & 26.87\nosig \\
+ GPT4~\cite{openai2023gpt4}                         & -0.143\nosig & -0.024\nosig & 0.030\nosig & 27.29\nosig & 26.71\nosig & 26.76\nosig \\
+ PromptistSFT~\cite{hao2022optimizing}                 & -0.140\nosig & -0.083\nosig & 0.010\nosig & 27.19\nosig & 26.60\nosig & 26.77\nosig \\
+ PR-All~\cite{dehghani2017learning}                       & 0.094\sigbase & 0.180\nosig & 0.233\sigbase & 27.51\sigbase & 26.91\sigbase & 26.95\sigbase \\
+ PR-Weighted                  & 0.083\sigbase & 0.164\nosig & 0.227\sigbase & 27.48\sigbase & 26.87\nosig & 26.97\sigbase \\
+ PR-Filter                    & 0.092\sigbase & 0.197\nosig & 0.241\sigbase & 27.48\sigbase & 26.91\sigbase & 26.98\sigbase \\
+ \textbf{CAPR}                         & \B 0.152\allsig & \B 0.213\nosig & \B 0.311\allsig & \B 27.56\allsig & \B 26.95\allsig & \B 27.04\allsig \\
\bottomrule
\end{tabular}
\end{table*}

\let\B\undefined
\let\allsig\undefined
\let\sigbase\undefined
\let\sigfilter\undefined
\let\nosig\undefined

\newcommand{\B}{\bfseries}

\newcommand{\allsig}{$^{*\dagger}$}
\newcommand{\sigbase}{$^{*}$\;\,}
\newcommand{\sigfilter}{$^{\dagger}$\;\,}
\newcommand{\nosig}{\;\;\,\,}

\begin{table*}
\centering
\small
\caption{Performance on an unseen and advanced system~(SDXL). ImageReward~\cite{xu2020neural} and HPSv2~\cite{wu2023humanv2} serve as evaluation models ($f$ in Eq.~(\ref{eq:evaluation})) and numbers are the average output scores.
$^{*}$ and $^{\dagger}$ separately indicate that performance is significantly better than SDXL and SDXL+PR-Filter at $p < 0.01$ level measured by ttest. CAPR effectively transfers to this unseen system.}
\label{tab:compare_on_sdxl}
\setlength{\tabcolsep}{3mm}
\begin{tabular}{l|ccc|ccc}
   	\toprule
    \multirow{2}{*}{Method} & \multicolumn{3}{c|}{ImageReward} & \multicolumn{3}{c}{HPSv2} \\
    & Anime & ConceptArt & Painting & Anime & ConceptArt & Painting \\
    \midrule
SDXL~\cite{podell2023sdxl}                         & 0.992\nosig & 0.903\nosig & 0.907\nosig & 28.37\nosig & 27.46\nosig & 27.52\nosig \\
+ GPT3.5~\cite{brown2020language}                       & 0.883\nosig & 0.762\nosig & 0.832\nosig & 28.26\nosig & 27.32\nosig & 27.50\nosig \\
+ GPT4~\cite{openai2023gpt4}                         & 0.831\nosig & 0.743\nosig & 0.794\nosig & 28.28\nosig & 27.34\nosig & 27.43\nosig \\
+ PromptistSFT~\cite{hao2022optimizing}                 & 0.785\nosig & 0.638\nosig & 0.688\nosig & 28.12\nosig & 27.19\nosig & 27.34\nosig \\
+ PR-All~\cite{dehghani2017learning}                       & 1.014\nosig & 0.926\sigbase & 0.948\sigbase & 28.43\sigbase & 27.51\sigbase & 27.61\sigbase \\
+ PR-Weighted                  & 1.008\nosig & 0.919\nosig & 0.941\sigbase & 28.44\sigbase & 27.52\sigbase & 27.62\sigbase \\
+ PR-Filter                    & 1.025\sigbase & 0.918\nosig & 0.947\sigbase & 28.44\sigbase & 27.53\sigbase & 27.61\sigbase \\
+ \textbf{CAPR}                         & \B 1.069\allsig & \B 0.949\allsig & \B 1.023\allsig & \B 28.50\allsig & \B 27.56\allsig & \B 27.68\allsig \\
\bottomrule
\end{tabular}
\end{table*}

\let\B\undefined
\let\allsig\undefined
\let\sigbase\undefined
\let\sigfilter\undefined
\let\nosig\undefined

\subsubsection{Baselines}

We compare CAPR against a comprehensive set of reformulation models, including:
\begin{itemize}
	\item \textbf{GPT3.5 \& 4}~\cite{brown2020language, openai2023gpt4}: They are used via APIs to reformulate prompts. We modify a popular prompt template from the web~\cite{bluelovers2023chatgpt} to guide them for this task. The prompt contains task descriptions and well-crafted prompt examples. 
	\item \textbf{PromptistSFT}~\cite{hao2022optimizing}: This reformulation model is trained on synthesized reformulation pairs. The researchers first crawl well-performing prompts from online websites where users share prompts. Then they use ChatGPT to rephrase these prompts to poor prompts. Finally, they train a language model to predict the original prompts from the poor prompts.
	\item \textbf{PR-All}~\cite{dehghani2017learning}: This is a traditional reformulation model. It utilizes a sequence-to-sequence transformer to predict the reformulated prompt. Compared to CAPR, it is trained on the same training data except that it does not introduce a condition mechanism. 
	\item \textbf{PR-Weighted}: Compared to PR-All, it resolves the data quality problem by weighting each training pair based on its quality. It uses the quality improvement measured by ImageReward~\cite{xu2023imagereward} to weight loss.
	\item \textbf{PR-Filter}: Compared to PR-All, it resolves the data quality problem by filtering out the low-quality pairs. The quality is measured as the improvement of ImageReward~\cite{xu2023imagereward} score. We tune and select the best filtering threshold. 
\end{itemize}
Note that except for GPT3.5 \& 4, the remaining baselines are all trained as a simple sequence-to-sequence translation model, as in previous reformulation models in web search scenarios~\cite{dehghani2017learning}.
They do not employ a condition in the model designs and simply view reformulation as a source-target translation task.

\subsubsection{Implementation Details} \mbox{}

CRM is initialized with TinyLlama~\cite{zhang2024tinyllama}, a 1.1B model trained on 2.5T tokens. Training lasts for $2$ epochs with AdamW~\cite{kingma2014adam} optimizer and a learning rate of $4 \times 10^{-5}$.
When tuning CCF, we construct a validation set of 100 prompts and set the inference steps of SD1.4 to $20$ steps for acceleration. We use the gp\_minimize function from skopt package~\cite{head_2021_5565057} to efficiently search the optimal CCF values within $50$ generation calls. 
For more details, please refer to our open-sourced code.

\subsection{Experimental Results}

In this section, we present the experimental results. The experiments are conducted on two text-to-image generation systems, namely SD1.4 and SDXL. SD1.4 is the model that corresponds to the training interaction log and thus has been seen by CAPR and baselines, while SDXL has not been seen by the reformulation models and thus can evaluate the model robustness. 

\subsubsection{Performance on the Seen System.}
The performance on SD1.4 is shown in Table~\ref{tab:compare_on_sd14}. Results demonstrate that CAPR substantially improves the generation quality and significantly outperforms all baselines. Specifically, we have the following observations:
\begin{itemize}
	\item Generic language models like GPT3.5 and GPT4 cannot improve the generation quality. After manually examining the results, we find that although both models try to mimic the formulation of well-crafted prompts, they tend to hallucinate and change the prompt meanings. Besides, although the output of GPT4 is more organized than GPT3.5 in terms of prompt structure, GPT4 adds a lot of modifier words like image style that misalign with user intention. This adversarially leads to worse results than GPT3.5. 
	\item According to the performance of PromptistSFT, training on synthetically generated refinement pairs results in limited effectiveness. In its training data, the target labels are well-crafted prompts crawled from the web, while the user inputs are simulated by rephrasing these prompts to everyday language with ChatGPT. Nevertheless, the distribution of such rephrased text is different from that of real users' input, resulting in an ineffective model.
	\item Training on users' reformulation data helps models learn to reformulate. PR-All, PR-Weighted, and PR-Filter all improve the generation quality. We also observe that PR-Weighted and PR-Filter perform similarly to PR-All. Although both models preprocess the training data to focus on training data with high quality, this preprocessing also results in limited training data size, which adversarially affects model training. 
	\item CAPR significantly outperforms the baselines. Compared with generic language models and PromptistSFT, CAPR is trained on real users' reformulation data and effectively learns useful reformulation strategies. Compared with PR-All/Weighted/Filter,  CAPR adopts a conditional reformulation framework that can better address the quality issue of user reformulation data.
\end{itemize}

\subsubsection{Performance on the Unseen System.}

Table~\ref{tab:compare_on_sdxl} shows the performance on SDXL~\cite{podell2023sdxl}, an advanced model not used during training. Results suggest that CAPR effectively transfers to this new model and significantly improves the generation performance. We have the following observations:
\begin{itemize}
	\item SDXL can evaluate the robustness of our reformulation models because it is substantially more advanced than SD1.4. According to Table~\ref{tab:compare_on_sd14} and Table~\ref{tab:compare_on_sdxl}, SDXL substantially improves the generation performance of SD1.4. Since the interaction log used for training is for SD1.4, SDXL can evaluate how reformulation models generalize to more advanced models. 
	\item CAPR still significantly improves the generation performance of SDXL and outperforms all baselines. This demonstrates that the benefits of CAPR are parallel to the advance of text-to-image generation system. 
\end{itemize}

\begin{figure}
    \centering
    \subfloat[Effects of ``Overall Quality''.]{
        \includegraphics[width=0.45\columnwidth]{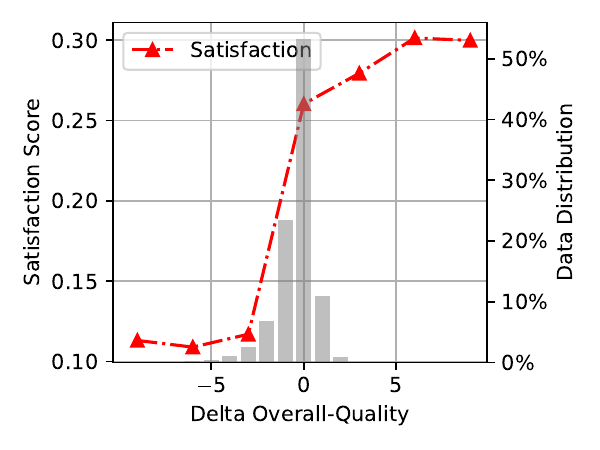}
        \label{fig:analysis_ccf_overall}
    }
    \hfill
    \subfloat[Effects of ``Similarity''.]{
        \includegraphics[width=0.45\columnwidth]{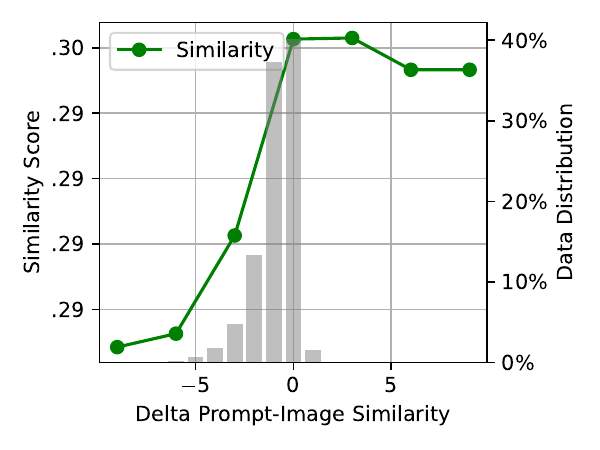}
        \label{fig:analysis_ccf_sim}
    }
    
    \subfloat[Effects of ``Aesthetic Quality''.]{
        \includegraphics[width=0.45\columnwidth]{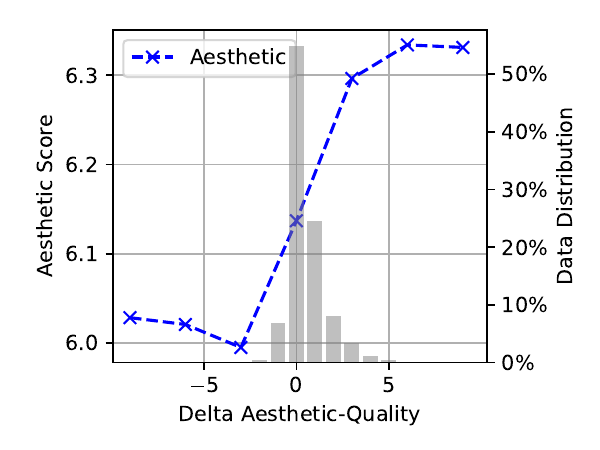}
        \label{fig:analysis_ccf_aesthetic}
    }
    \hfill
	\subfloat[Effects of ``Prompt Length''.]{
        \includegraphics[width=0.45\columnwidth]{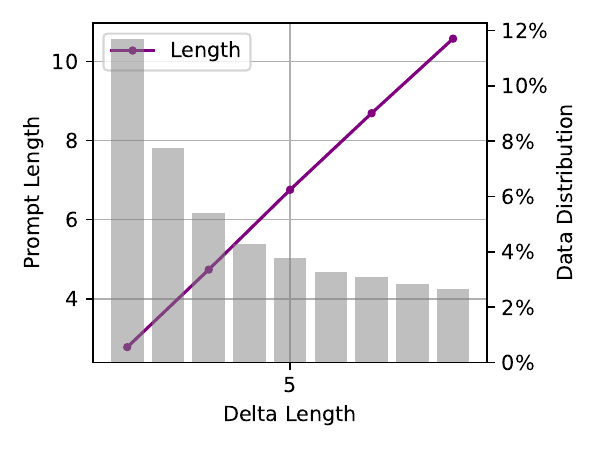}
        \label{fig:analysis_ccf_length}
    }
    \caption{
    Effects of CCF conditions for CRM. X-axis is the input expected performance improvement, as formulated in Eq.~(\ref{eq:ccf_reparameterization}). Lines show the evaluation output quality, and bars show the training data distribution. Results suggest that CRM can be effectively controled by the input condition.
    }
    \label{fig:analysis_ccf_metric}
    
\end{figure}


\subsection{Model Analysis}

The key of CAPR is to guide the reformulation behaviour using configurable feature conditions (CCF). In this section, we deeply analyze how different conditions control the CAPR performance. 

The methodology is to tune each factor separately and observe the generation quality. 
We use the ``Painting'' dataset for analysis. We tune the expected reformulation quality, as formulated in Eq.~(\ref{eq:ccf_reparameterization}), i.e., requiring CAPR to improve a factor by a specified delta value ($\delta$ in Eq.~(\ref{eq:ccf_reparameterization})). The optimal delta values learned from the validation data (described in Section~\ref{sec:tune_expected_ccf}) are $9$ for overall quality, $9$ for aesthetic quality, $0$ for prompt-image similarity, and $5$ for prompt length. 
To analyze the effects of each factor, we freeze other factors to eliminate their influence: keeping the delta values for prompt length to $5$ and other factors to $0$.
In the following, we discuss how CAPR follows each condition factor.

\newlength{\myimagewidth}
\setlength{\myimagewidth}{0.17\linewidth} 

\begin{table}[t]
\centering
\scriptsize
\caption{Prompt reformulation examples. Column one details user prompts and the reformulated prompts by PR-Filter and CAPR. The next three columns show images generated by SD1.4 from user inputs or reformulated prompts. CAPR substantially enhances the image quality.}
\label{tab:prompt_case_study}
\setlength{\tabcolsep}{1pt}
\begin{tabular}{m{0.45\linewidth}ccc}
\toprule
\textbf{Input Prompt} & \textbf{User Input} & \textbf{PR-Filter} & \textbf{CAPR} \\ 
\midrule
\parbox[b]{\linewidth}{\textbf{User Input}: A monkey is pictured acting as a DJ. \\ 
\textbf{PR-Filter}: A monkey DJ. \\
\textbf{CAPR}: A monkey, wearing headphones, A monky is pecturing as a dj., digital art, artstation, by greg rutkowski}
& 
\raisebox{-.5\height}{\includegraphics[width=\myimagewidth]{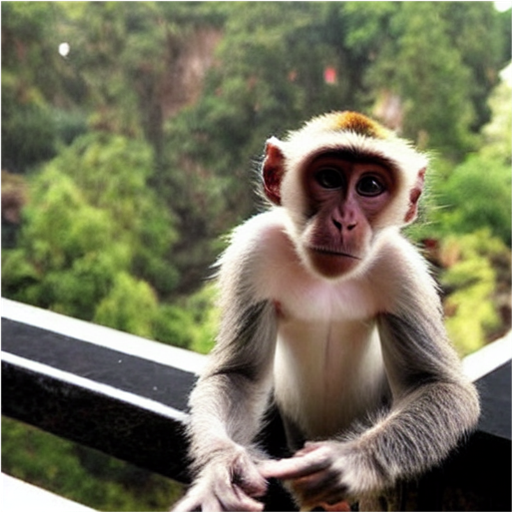}} 
& 
\raisebox{-.5\height}{\includegraphics[width=\myimagewidth]{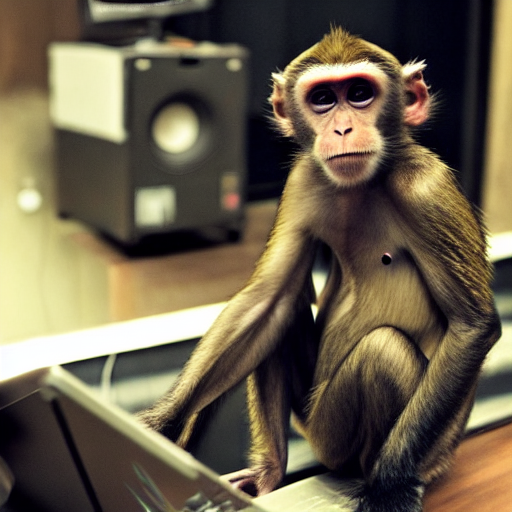}}
& 
\raisebox{-.5\height}{\includegraphics[width=\myimagewidth]{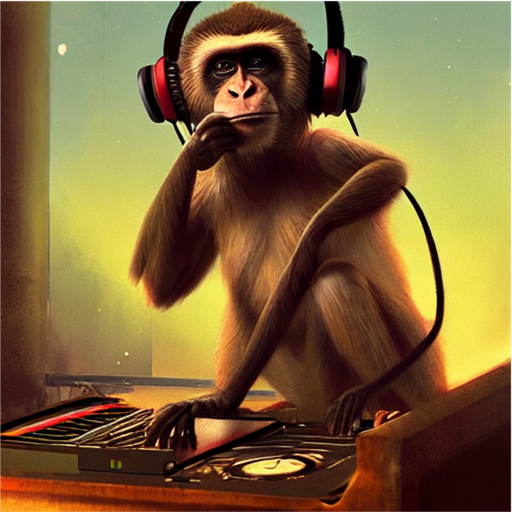}}
\\
\midrule
\parbox[b]{\linewidth}{\textbf{User Input}: Milt Kahl's sketch of Cecil Turtle. \\ 
\textbf{PR-Filter}: Milt Kahl sketch of a turtle. \\
\textbf{CAPR}: Milt Kahl's sketch cecil turtle. detailed, high quality, digital painting, fantasy, artwork, in the style of Cecilia Turtles}
& 
\raisebox{-.5\height}{\includegraphics[width=\myimagewidth]{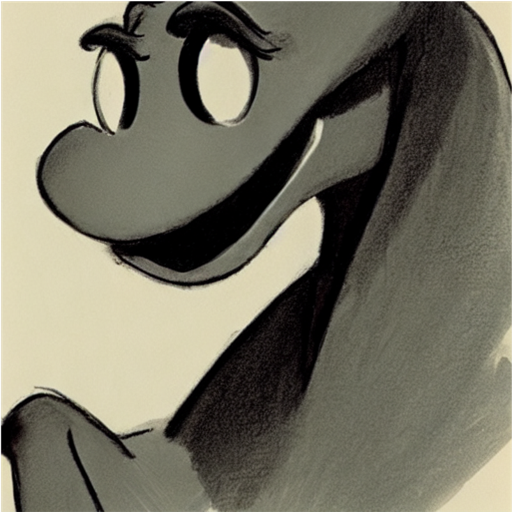}} 
&
\raisebox{-.5\height}{\includegraphics[width=\myimagewidth]{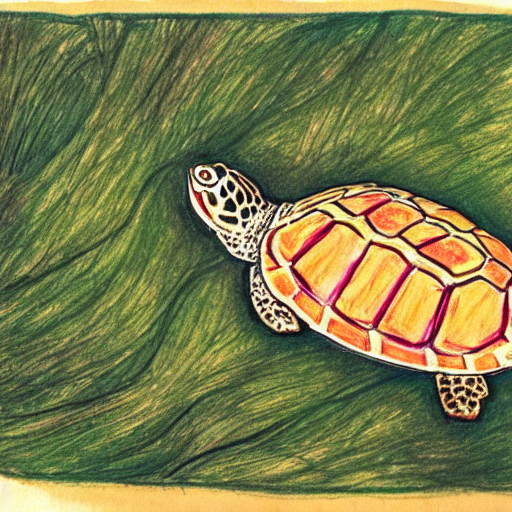}}
& 
\raisebox{-.5\height}{\includegraphics[width=\myimagewidth]{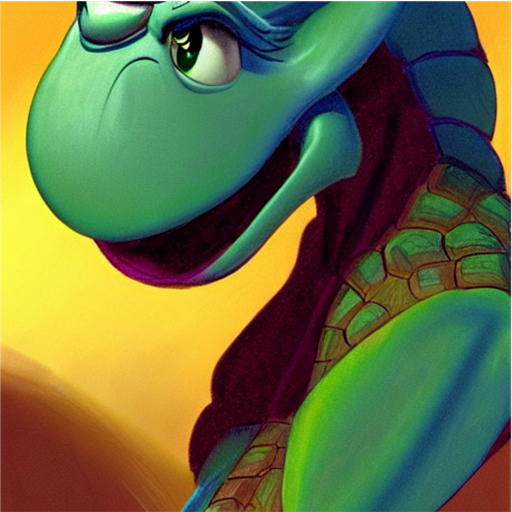}}
\\
\midrule
\parbox[b]{\linewidth}{\textbf{User Input}: An elephant carrying a house on its back. \\ 
\textbf{PR-Filter}: An elephantine elephantin carrying a human on its shoulders. \\
\textbf{CAPR}: An elephant carrying a House on its Back. Fantasy, digital painting, HD, 4k, detailed, artwork}
& 
\raisebox{-.5\height}{\includegraphics[width=\myimagewidth]{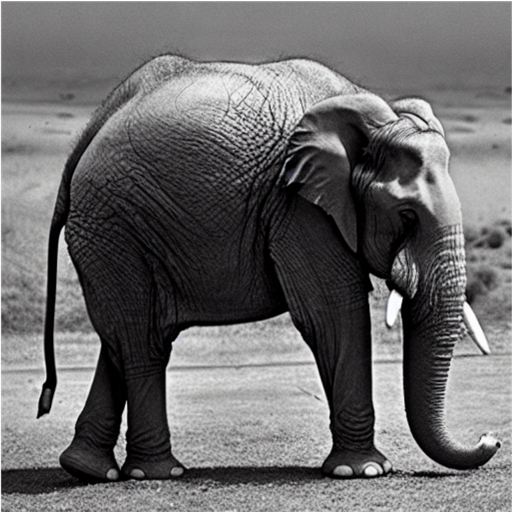}} 
&
\raisebox{-.5\height}{\includegraphics[width=\myimagewidth]{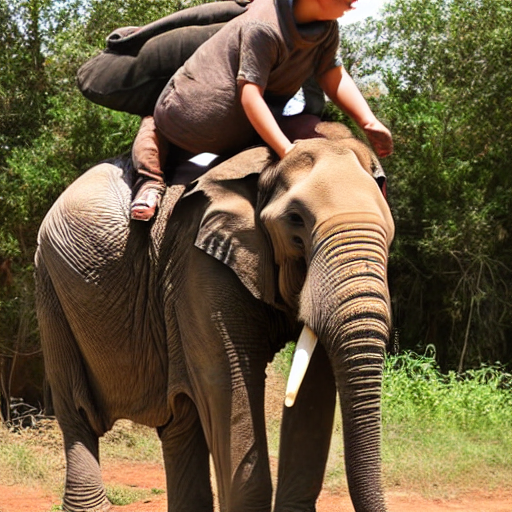}}
& 
\raisebox{-.5\height}{\includegraphics[width=\myimagewidth]{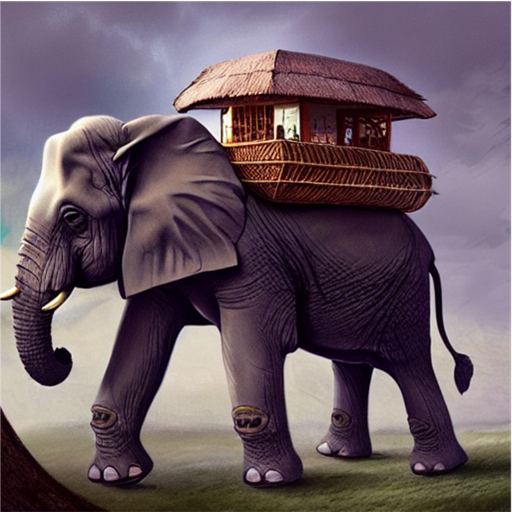}}
\\
\midrule
\parbox[b]{\linewidth}{\textbf{User Input}: The interior of a spaceship orbiting alpha centauri. \\ 
\textbf{PR-Filter}: The interior of an alien spaceship. \\
\textbf{CAPR}: The interior of spaceship of a fantasy setting, highly detailed, digital painting, artstation, concept art, illustration}
& 
\raisebox{-.5\height}{\includegraphics[width=\myimagewidth]{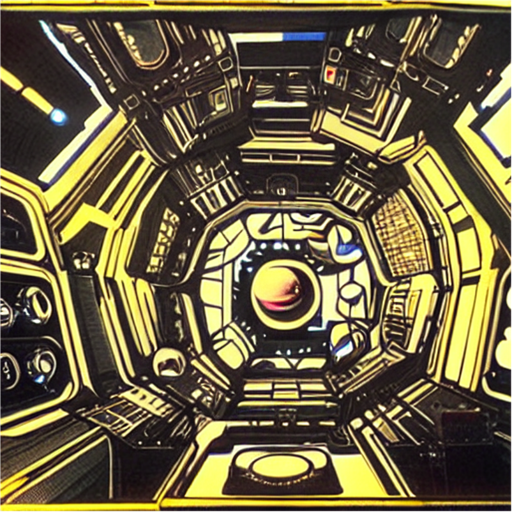}} 
& 
\raisebox{-.5\height}{\includegraphics[width=\myimagewidth]{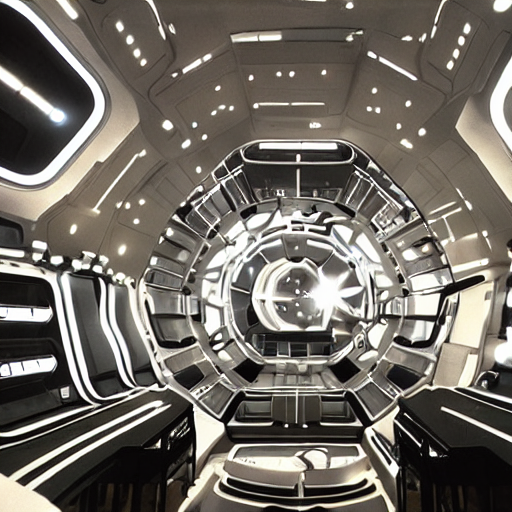}}
& 
\raisebox{-.5\height}{\includegraphics[width=\myimagewidth]{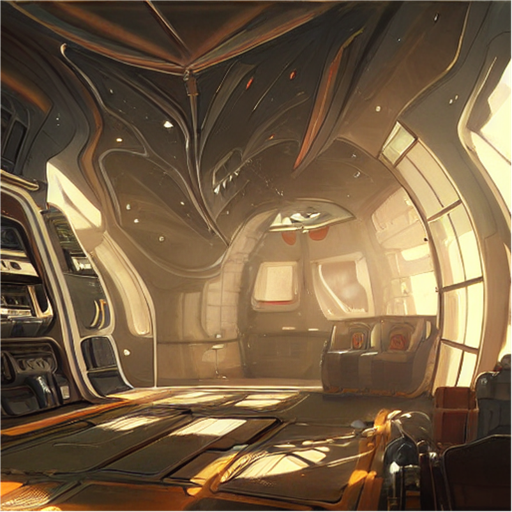}}
\\
\bottomrule
\end{tabular}
\end{table}

\let\myimagewidth\undefined

As shown in Figure~\ref{fig:analysis_ccf_metric}, CAPR can follow the specified condition well and even extrapolate beyond the training data distribution. Specifically, when we gradually increase the delta values, the corresponding metric scores improve. This demonstrates that CCF can control the behaviour of CAPR.
In Figure~\ref{fig:analysis_ccf_metric}, we also depict the distribution of delta scores for each CCF factor in the training data (i.e., the grey bars in the figures). We can see that most user-reformulated prompts have near-zero delta scores in terms of different image quality measurements. This low-quality training data is primarily caused by the fact that users' capacity is relatively stable during one session, as discussed in Section~\ref{sec:analysis_prompt_reform}. Nevertheless, CAPR can extrapolate beyond the limits of the training. For example, although few training pairs improve the overall quality more than $2$, as shown in Figure~\ref{fig:analysis_ccf_overall}, CAPR can still improve user satisfaction when the condition is increased from $2$ to $6$.

\subsection{Case Studies}

Table~\ref{tab:prompt_case_study} includes four reformulation examples. 
We can see that PR-Filter only slightly rephrases user inputs and even hallucinates in the third case. This stems from the training data where human-generated reformulation pairs are alike, as discussed in Section~\ref{sec:analysis_prompt_reform}.
Instead, CAPR reformulates prompts to keyword-enriched ones with artist names and stylistic elements, which have been demonstrated to be favored by text-to-image generation systems~\cite{oppenlaender2022taxonomy, witteveen2022investigating}. 
For instance, in the first example about a monkey DJ, CAPR adds the ``headphones'' detail and the artist name ``greg rutkowski''. 
These cases demonstrate that the conditional framework is the key to learning prompt reformulation from interaction logs.

\section{Conclusion}

Text-to-image generation systems have increasingly become a milestone in digital art creation. Yet, their effectiveness is closely tied to the quality of the prompts provided by users, a task that often presents significant challenges to the average user. In this paper, we leverage user interaction logs as a valuable resource for training an automatic prompt reformulation model. Our investigation reveals a distinctive aspect of prompt reformulation in text-to-image systems: it relies heavily on the user's intrinsic ability to craft effective prompts rather than on the system's feedback.
To address this unique challenge, we introduce the Capability-aware Prompt Reformulation (CAPR) framework, a pioneering solution for training on interaction logs. CAPR can adapt to various user capabilities and simulate high-quality reformulation during inference. Extensive experiments demonstrate the effectiveness of CAPR, highlighting its significant improvements over existing baselines and transferability to unseen systems.

\begin{acks}
This work is supported by Quan Cheng Laboratory (Grant No. QCLZD202301).
\end{acks}


\newpage
\bibliographystyle{ACM-Reference-Format}
\balance
\bibliography{references}

\newpage

%
%
%

\end{document}